\documentclass[letterpaper, 10 pt, conference]{ieeeconf}  

\IEEEoverridecommandlockouts                              

\usepackage{cite}
\usepackage{amsmath,amssymb,amsfonts}
\usepackage{graphicx}
\usepackage{subcaption}
\usepackage{textcomp}
\usepackage{xcolor}
\usepackage{float}
\usepackage{algorithm}
\usepackage{algpseudocode}

\makeatletter
\let\NAT@parse\undefined
\makeatother
\usepackage{hyperref}

\hypersetup{
    colorlinks,
    linkcolor={red!50!black},
    citecolor={red!50!black},
    urlcolor={blue!80!black}
}
\title{\LARGE \bf
Revisiting Proprioceptive Sensing for Articulated Object Manipulation
}

\author{Thomas Lips$^{* \dagger}$ and Francis wyffels$^{\dagger}$
\thanks{$^{\dagger}$ AI and Robotics Lab, IDLab-AIRO, Ghent University -- imec}%
\thanks{$^{*}$ corresponding author: \url{Thomas.Lips@ugent.be}}
}

\begin{document}

\maketitle

\begin{abstract}
Robots that assist humans will need to interact with articulated objects such as cabinets or microwaves. Early work on creating systems for doing so used proprioceptive sensing to estimate joint mechanisms during contact. However, nowadays, almost all systems use only vision and no longer consider proprioceptive information during contact. We believe that proprioceptive information during contact is a valuable source of information and did not find clear motivation for not using it in the literature. Therefore, in this paper, we create a system that, starting from a given grasp, uses proprioceptive sensing to open cabinets with a position-controlled robot and a parallel gripper. We perform a qualitative evaluation of this system, where we find that slip between the gripper and handle limits the performance. Nonetheless, we find that the system already performs quite well. This poses the question: should we make more use of proprioceptive information during contact in articulated object manipulation systems, or is it not worth the added complexity, and can we manage with vision alone? We do not have an answer to this question, but we hope to spark some discussion on the matter. The codebase and videos of the system are available \href{https://tlpss.github.io/revisiting-proprioception-for-articulated-manipulation/}{here}.

\end{abstract}

\section{INTRODUCTION}

Our living environments contain many articulated objects, including storage furniture such as cabinets and drawers or appliances like dishwashers and microwaves. Interacting with such objects will be a crucial skill of assistive robots and has hence been of great interest in robotic manipulation research.

Some of the earliest works on articulated object manipulation are~\cite{jain2010pulling} and~\cite{sturm2011probabilistic}. These works perform explicit estimation of the joint parameters (type of the joint, axis of rotation/translation, and joint configuration) based on a sequence of part poses. The poses are obtained from the end-effector (proprioceptive sensing) under the assumption of a \textit{firm grasp} (i.e. rigid connection between handle and gripper), or from fiducial markers~\cite{sturm2011probabilistic}. They use a hook to grasp the handle of the articulated objects at manually specified poses and use a compliant controller to overcome inaccuracies in the joint estimations to avoid exerting large forces. \cite{jain2010pulling} obtains an impressive 37/40 success rate when tested on several articulated objects.

Other researchers have also estimated the articulation parameters either directly from a series of images directly~\cite{jain2021screwnet}, or by first tracking the poses of the parts and then estimating the parameters from this sequence of poses~\cite{heppert2022category}. Yet other work has learned to detect articulated objects and determine their joint parameters from a single pair of stereo RGB images~\cite{heppert2023carto}.

\begin{figure}
\centering
\begin{subfigure}{.5\linewidth}
  \centering
  \includegraphics[width=0.95\linewidth]{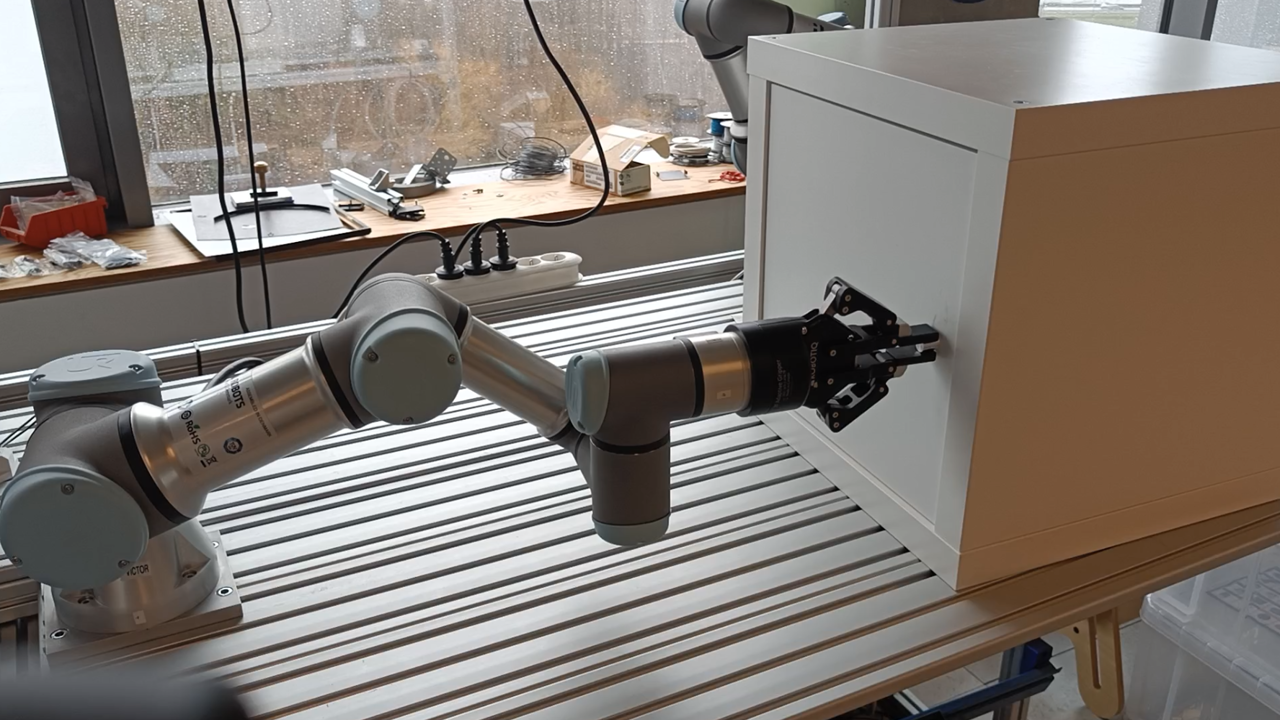}
\end{subfigure}%
\begin{subfigure}{.5\linewidth}
  \centering
  \includegraphics[width=0.95\linewidth]{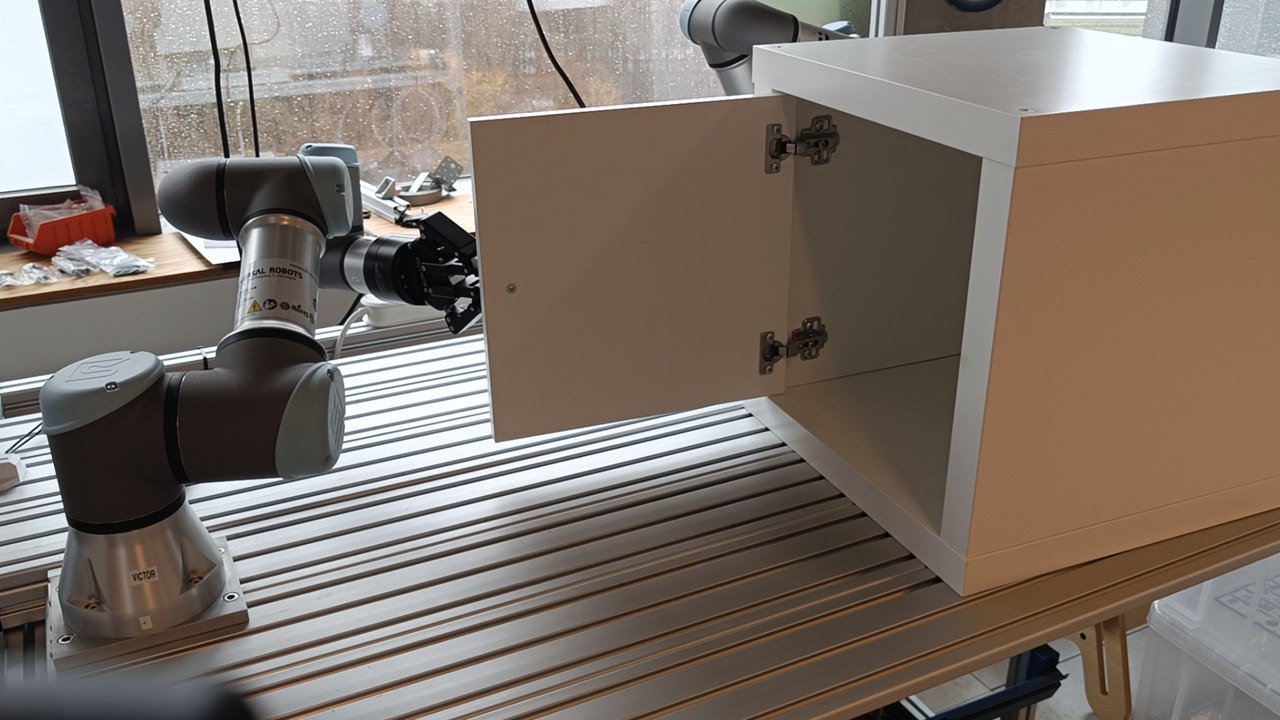}
\end{subfigure}
\caption{We create a system to open articulated objects using only the proprioceptive information during contact. The system can open various articulated objects. We find that the main limitation is the occurrence of slip between the handle and gripper as can be observed in the different orientations of the gripper w.r.t. handle in the images above.}
\label{fig:first-page-example}
\end{figure}

Another line of work has focused on learning affordances for actions instead of explicitly determining joint parameters~\cite{mo2021where2act,xu2022UMP,eisner2022flowbot3d}. These still perform separate grasp generation but then use closed-loop affordance estimation to determine appropriate actions for the robot to open the articulated objects. So far, all work in this direction determines the affordances based on a single observation and does not adapt at inference to correct wrong predictions \cite{eisner2022flowbot3d}. 

Almost all papers mentioned so far use the PartNet-Mobility dataset~\cite{xiang2020partnet-mobility} to obtain the required training data. Many use suction cups to limit the complexity of grasping the articulated object~\cite{xu2022UMP,eisner2022flowbot3d}. Even then, determining appropriate grasping poses is challenging and is often reported as a major failure mode~\cite{xu2022UMP, eisner2022flowbot3d}.
Most works use a force-controlled robot to manipulate the articulated objects and a compliant\footnote{As in~\cite{calanca2015reviewcompliant}, we categorize compliant control as control schemes that shape the relation between positions (or velocities) and external forces} low-level control scheme such as Impedance Control~\cite{impedance} or Operation Space Control~\cite{khatib-OSC} to account for uncertainties in the joint parameters. 

There are also more end-to-end works that aim to use Reinforcement Learning and Imitation Learning to open articulated objects~\cite{gupta2019relay,ebert2021bridgedataset,shridhar2022peract}. These methods should be capable of handling the long tail that characterizes most category-level skills but they tend to require larger amounts of data and have so far not shown the same level of generalization on articulated object manipulation as the more task-specific methods that were discussed before.

A clear trend in recent work is to rely more on vision and not use proprioceptive information obtained during contact. However, to the best of our knowledge, this is not thoroughly motivated in the literature. Furthermore, proprioceptive information is naturally invariant to many of the typical varieties found in articulated objects and their environments, including materials, lighting, and certain aspects of the geometry.
Therefore, in this paper, we create a system that uses proprioceptive sensing to open articulated objects. Compared to~\cite{jain2010pulling}, we use a position-controlled robot and hence switch to an admittance control~\cite{impedance} scheme to make the end-effector compliant. We also use a parallel-position gripper, as this is more generic than task-specific end-effectors or suction cups. 

We describe the system in more detail in section~\ref{section:method}. In section~\ref{section:analysis} we qualitatively analyze the performance of our system on three articulated objects: 2 Ikea KALLAX cabinets and an oven. Based on this analysis, we then formulate the central question of this paper in section~\ref{section:to-use-or-not}: \textit{should we use proprioceptive information, or not?}. Finally, we make some suggestions to improve the evaluations of articulated object manipulation in section~\ref{section:discussion-evaluation}.

To summarize, our contributions are as follows:
\begin{itemize}
    \item We combine previous work to implement a system that can open articulated objects with proprioceptive sensing. We do this with a position-controlled robot and a parallel gripper. Our qualitative analysis shows that the system is capable of opening various articulated objects.
    \item We find that slip between the gripper and handle can lead to failures. Based on this observation we pose the question if the added complexity of dealing with this slip is worth the gains of using proprioceptive sensing.
    \item Finally, we formulate some suggestions on how to improve the evaluation of articulated object manipulation systems based on other failure modes that we encountered during the analysis.
\end{itemize}








\section{METHOD}
In this section, we describe how we open articulated objects using proprioceptive information. We use a position-controlled UR3e robot with a built-in force-torque sensor and a Robotiq 2F-85 parallel gripper. 




A high-level overview of our system is given in Algorithm~\ref{algorithm:overview}. In the following sections, we discuss the compliant controller and articulation estimation method in more detail. Note that we manually determine the grasp pose to limit the scope of this work. 

\begin{algorithm}
\caption{High-level overview of the system}
\label{algorithm:overview}
\begin{algorithmic}[1]

\State Enable the compliant controller \Comment{See section~\ref{section:compliant-control}}
\State Manually determine grasp pose
\State set the initial joint estimation to a prismatic joint in the -Z direction of the gripper
\While{cabinet not opened}
\Repeat{ $N$ times}
\State move  gripper along the current joint estimation 
\State collect the gripper pose $X^t$

\Until{}
\State obtain a new joint estimation from the sequence of previous gripper poses $\{X^{0:t}\}$ \Comment{see section~\ref{section:factor-graph}} 
\EndWhile
\end{algorithmic}
\end{algorithm}

\label{section:method}

\begin{figure*}
\centering
\begin{subfigure}{.24\linewidth}
  \centering
  \includegraphics[width=0.95\linewidth]{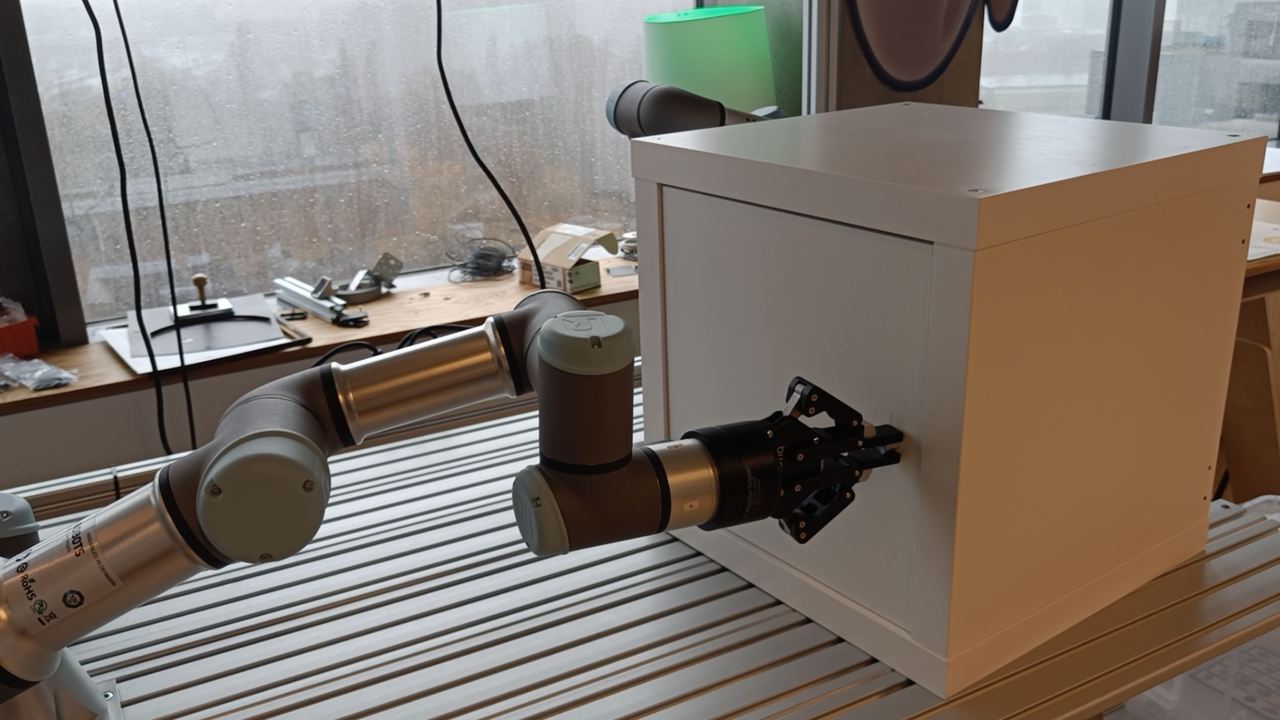}
  \includegraphics[width=0.95\linewidth]{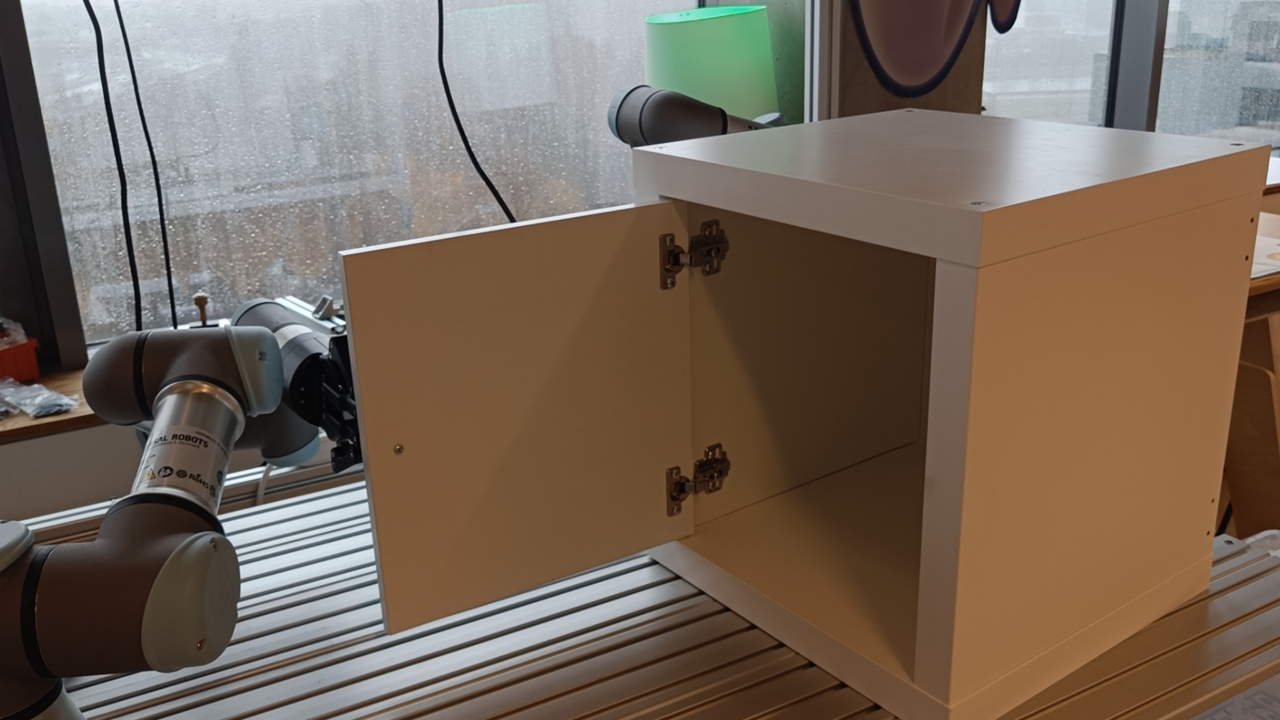}
    \caption{}
    \label{fig:examples-a}

\end{subfigure}
\begin{subfigure}{.24\linewidth}
  \centering
  \includegraphics[width=0.95\linewidth]{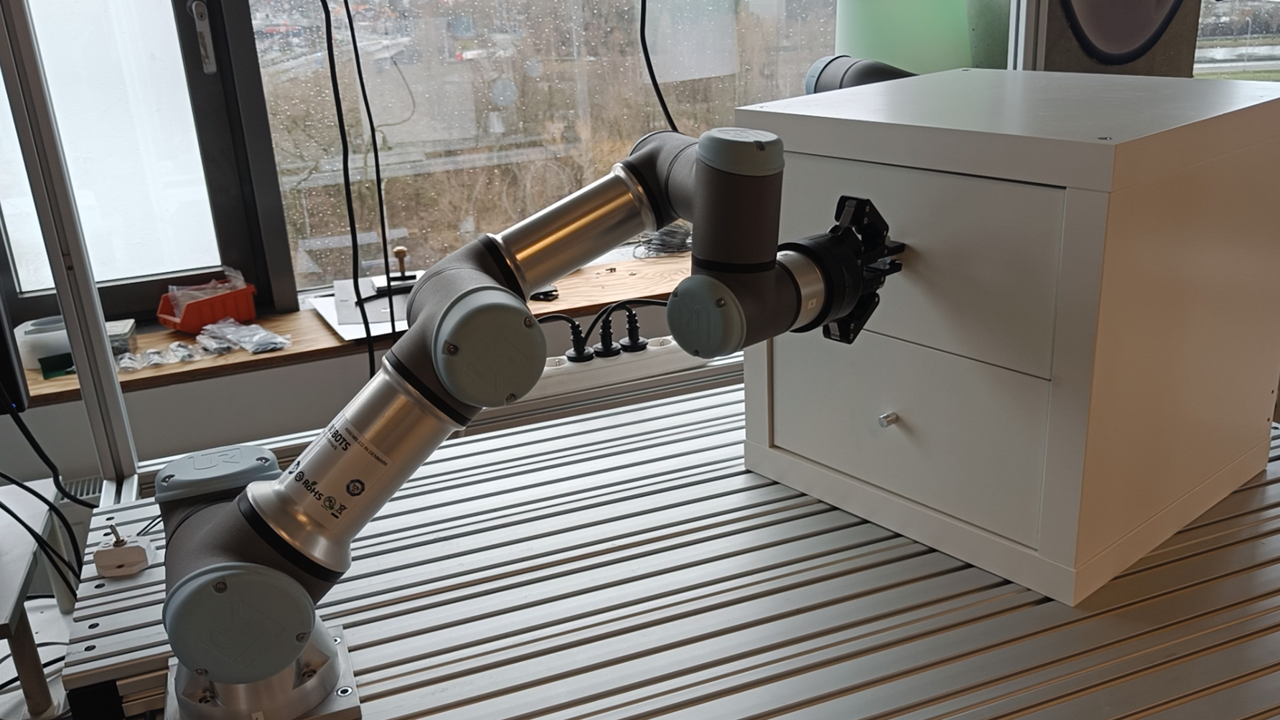}
  \includegraphics[width=0.95\linewidth]{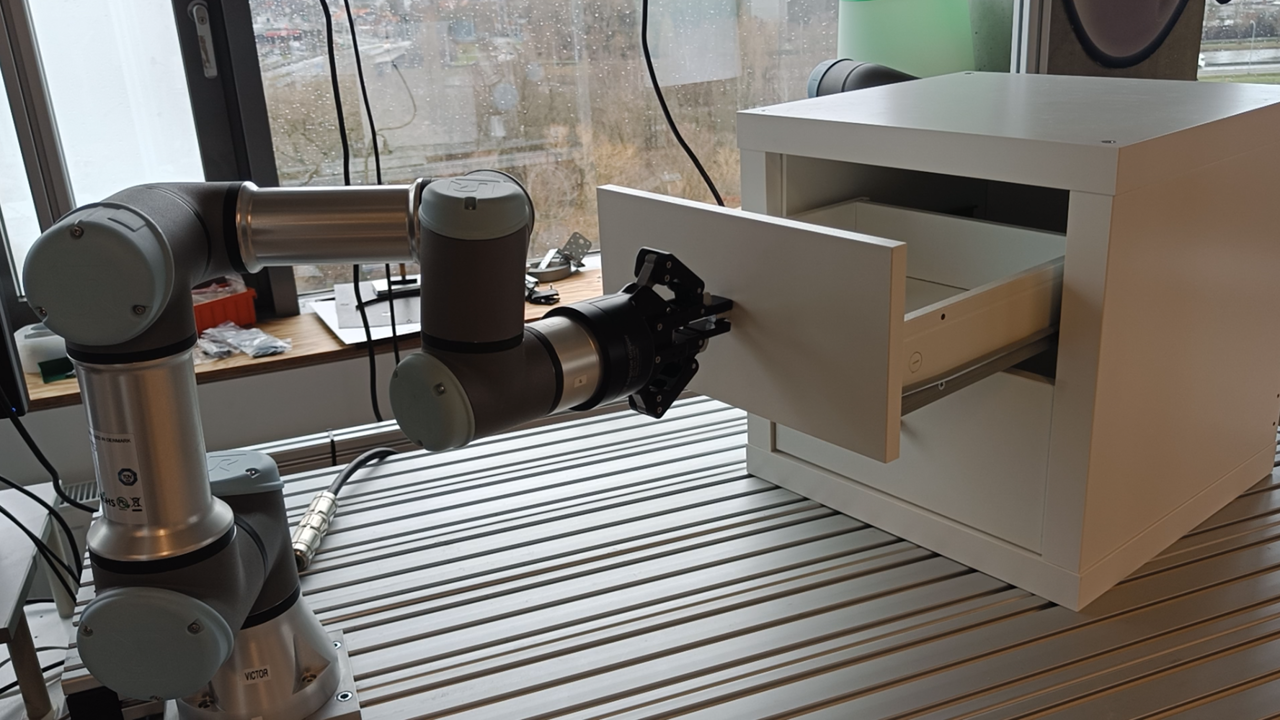}
    \caption{}
    \label{fig:examples-b}

\end{subfigure}\begin{subfigure}{.24\textwidth}
  \centering
  \includegraphics[width=0.95\linewidth]{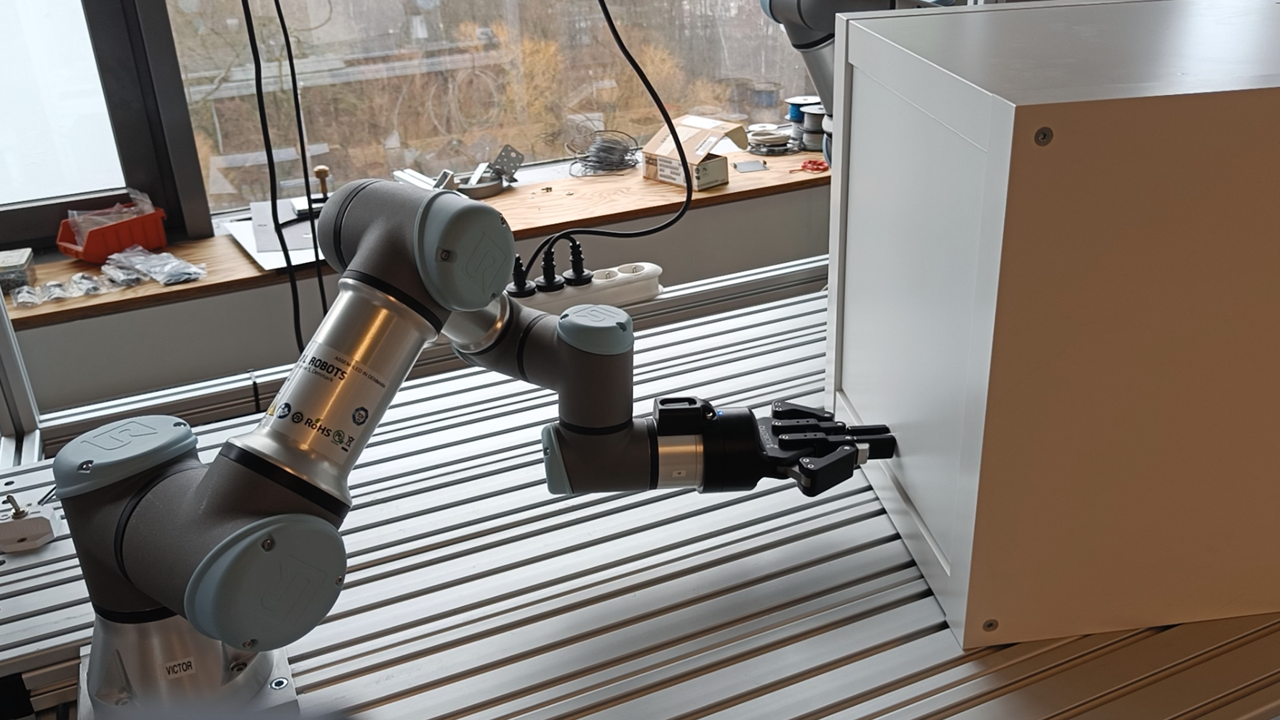}
    \includegraphics[width=0.95\linewidth]{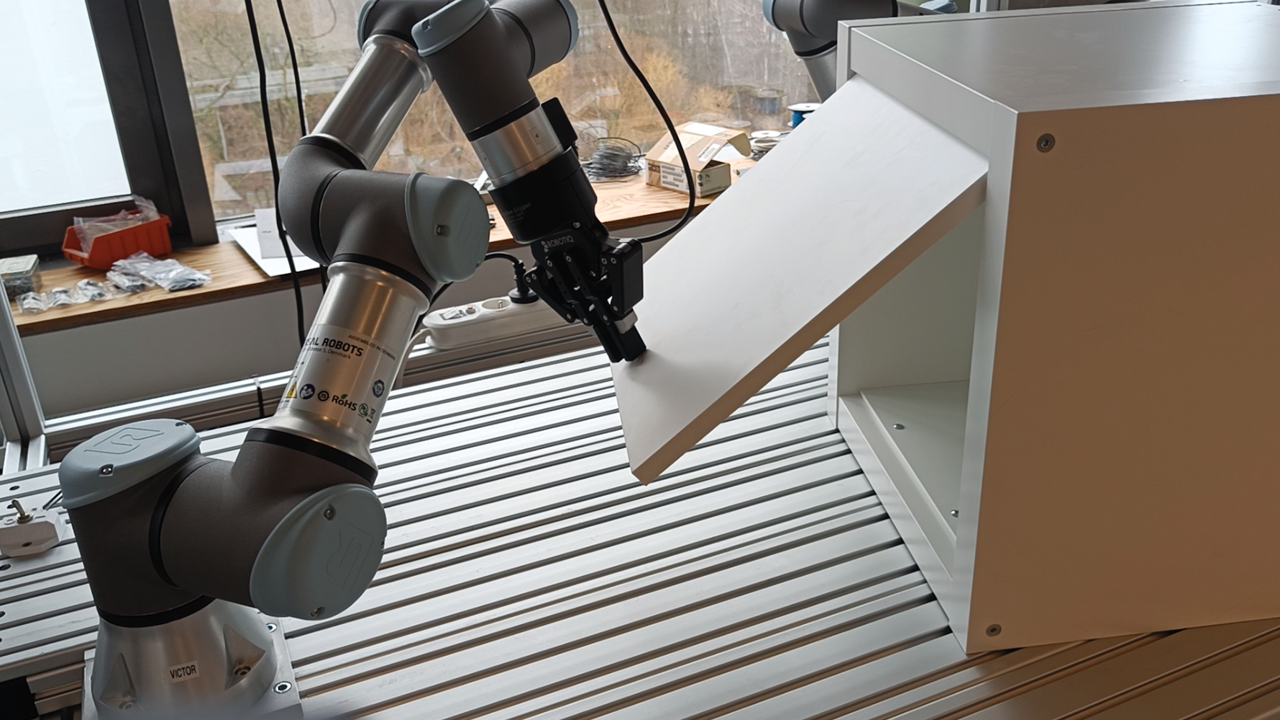}
    \caption{}
    \label{fig:examples-c}
    
\end{subfigure}%
\begin{subfigure}{.24\linewidth}
  \centering
  \includegraphics[width=0.95\linewidth]{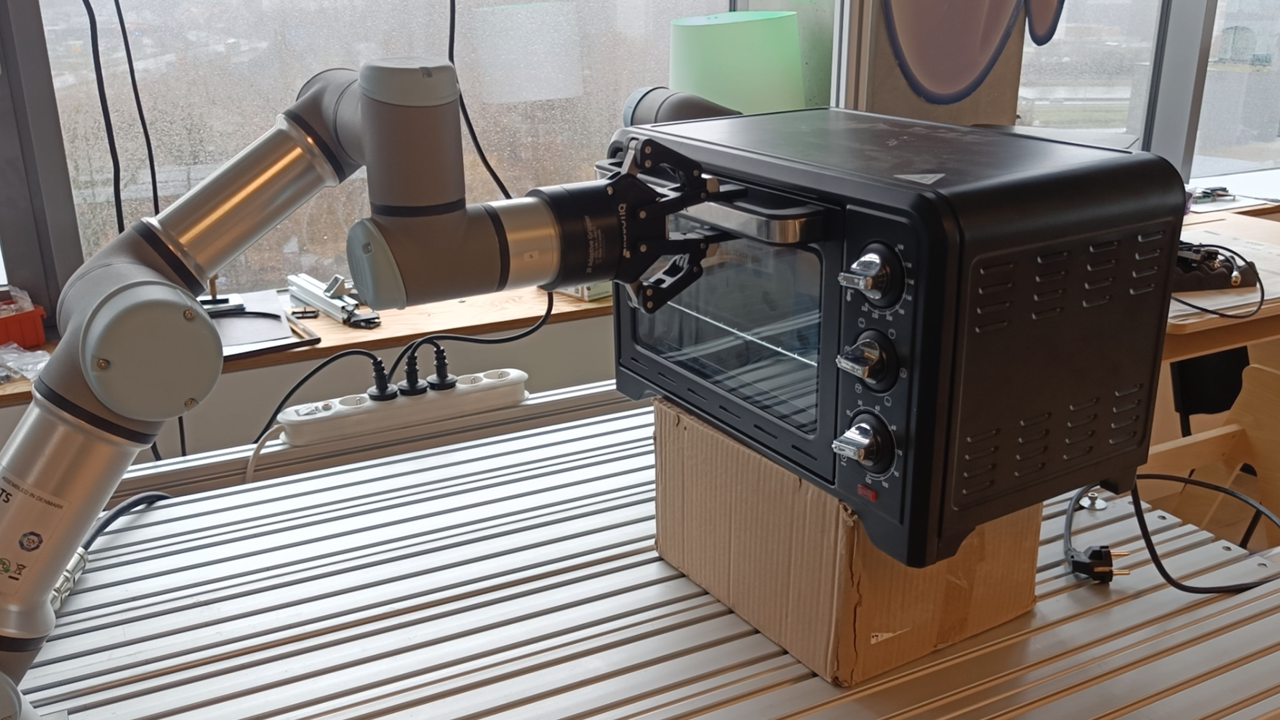}
  \includegraphics[width=0.95\linewidth]{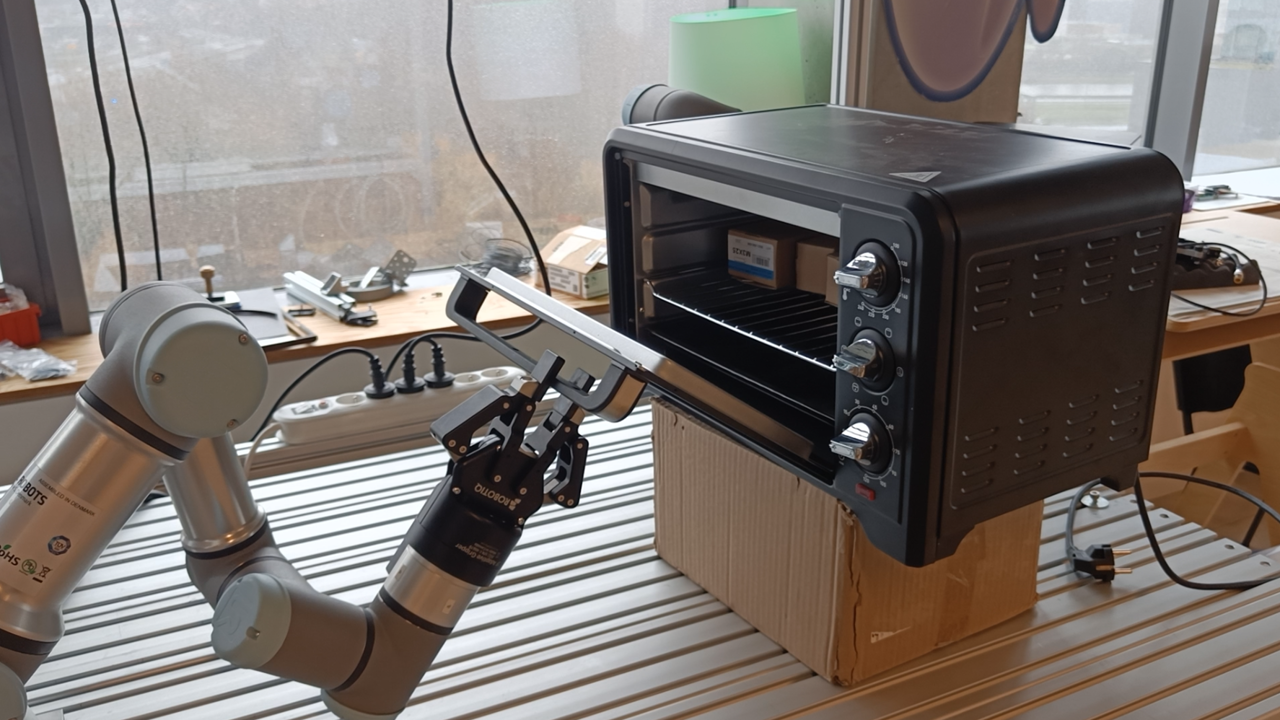}
    \caption{}
    \label{fig:examples-d}

\end{subfigure}

\caption{Additional examples of our system opening articulated objects. (a) and (b) are successful examples of opening a cabinet and drawer respectively. (c) shows how a fixed grasp limits the workspace of the robot such that it cannot fully open the cabinet. (d) shows how a fixed grasp can lead to collisions with the environment.}
\label{fig:test}

\end{figure*}
\subsection{Compliant controller}
\label{section:compliant-control}
To make the position-controlled UR3e robot compliant and hence capable of overcoming errors in the articulation estimation without applying excessive forces on the articulated object, we use an admittance control scheme.
With admittance control, we specify how the reference pose $X_\mathrm{r}$ of the end-effector should be adapted under external forces, thus making the robot compliant. The relation between the gravity-compensated wrench on the end-effector $W_{\mathrm{ext}}$ and the deviation from the reference trajectory $X_\mathrm{e}$ is shaped as follows:
\begin{equation}
     W_{\mathrm{ext}} = K X_\mathrm{e} + B\dot{X_\mathrm{e}} + M \ddot{X_\mathrm{e}},
\end{equation}
 where K, B and M are the stiffness, damping and mass matrix respectively~\cite{beltran2020learningforcecontrol}. The dimensions are often decoupled by setting the off-diagonal elements of the aforementioned matrices to zero. 
The desired pose for the robot end-effector is then determined as $X_d = X_r + X_e$.

A straightforward implementation of this scheme resulted in oscillatory and unstable behavior for stiff contacts (such as grasping cabinet handles) and low stiffness (which is desired to reduce forces applied on the gripper and hence avoid slip with inaccurate joint estimations). This is a known issue with low-stiffness admittance control on high-gain position-controlled robots such as the UR e-series~\cite{calanca2015reviewcompliant}. We, therefore, resorted to a more advanced implementation from~\cite{FZI-compliance}. We empirically set the translational stiffness to 200 N/m in the Z-direction of the gripper and 50 N/m in the X and Y directions. The rotational stiffness was set to 2 Nms/radian for all dimensions. Other parameters in the implementation were set to their default values.

\subsection{Articulation Estimation}
In this paper, we only consider joint mechanisms with a single degree of freedom as this is the most common case for furniture or appliances that can be \textit{opened} and \textit{closed}.
To estimate the joint parameters from the proprioceptive information, we used the method from Heppert et Al.~\cite{heppert2022category}, as the authors found their method, using a  Factor Graph formulation, to perform better than the probabilistic model that was used in~\cite{sturm2011probabilistic}. It also unifies the representation of revolute and prismatic joints as special cases of helical joints, which can be represented as a twist\footnote{We refer to~\cite{lynch2017modernrobotics} for an excellent introduction into the geometric interpretation of spatial algebra, that is used in this and many other works on articulated object manipulation.} $V \in \mathbb{R}
^6$.  The articulation estimation method takes in a sequence of part poses $\{X_{\mathrm{part}}^i\}$. These can be of any fixed frame on the moving part attached to the joint, e.g. a frame on the handle. Under the assumption that no slip occurs between the gripper and the handle, the poses of the gripper can also be used, which allows for estimating the joint parameters from the proprioceptive information.

The articulation estimation returns a joint twist estimation $\hat{V}$ and the joint configuration ${q_i}$ (that describes the configuration single DOF of the joint) for all poses. The twist can then be used to determine the pose of the part frame for any joint configuration by taking the matrix exponential of the skew-symmetric matrix of the twist~\cite{lynch2017modernrobotics}:
\begin{equation}
    X_{\mathrm{part}} = \exp(q [\hat{V}]) \in SE(3).
\end{equation}
This relation can then be used to open the articulated object. We set the initial joint estimation to a prismatic joint in the -Z direction of the initial grasp frame, as in \cite{jain2010pulling}.

To anticipate on slip between the gripper and handle, we increase the variance of the noise used in the factor graph model for the observed part poses compared to the original implementation. We also pre-compile all factor graphs to increase the speed of the estimations during manipulation. 

\label{section:factor-graph}

\subsection{Evaluation}
To evaluate the system, we selected 2 Ikea KALLAX\footnote{\url{https://www.ikea.com/be/en/cat/kallax-series-27534/}} cabinets, as they are widely available. One has a rotating door and one has 2 drawers. We also use an oven that is used in the lab for reflow soldering. All three items are relatively small to make sure they fit in the robot's workspace.

We do not report quantitative measures as we did not perform enough experiments nor had enough diverse cabinets for them to have statistical significance. Rather we provide a qualitative evaluation of the system and focus in particular on the failure modes as these are the most interesting in our opinion.


\section{RESULTS \& DISCUSSION}
In this section, we first perform an analysis of the system described in the previous section. Based on this analysis, we then pose the question if systems for articulated object manipulation should make more use of proprioception and formulate some suggestions on how to improve the evaluation of articulated object manipulation systems.

\subsection{Qualitative Analysis of the system}
\label{section:analysis}
We find that in general, the system is capable of opening the articulated objects, irrespective of the relative pose of the object and/or the presence of other objects in the environment. Initial joint estimations are sometimes largely different from the actual joint parameters but are usually good enough to continue opening the articulated object, which allows for collecting additional proprioceptive information and improves the articulation estimation. Some successful experiments can be seen in  Figures \ref{fig:examples-a} and \ref{fig:examples-b}. 

In the following paragraphs, we discuss three aspects of the system in more detail.

\subsubsection{slip}
As expected, due to inaccurate joint estimations that are used to determine the next target pose for the gripper, slip occurs between the handle and the gripper.  We found that this slip mostly affects the orientation of the grasp pose. The extent to which this slip occurs also largely depends on the shape of the handle, where handles with rounded sides such as in figure \ref{fig:examples-a} are more prone to slip than rectangular handles as in figure \ref{fig:examples-d}, as these provide a larger contact surface. We found that slip, even though it results in estimation errors for the joint parameters, not always results in a failure to open the articulated object. The slip is usually limited and hence results in limited changes to the perpendicular orientation of the gripper to the surface of the moving parts. An example of a successful interaction despite the occurrence of slip can be seen in Figure \ref{fig:first-page-example}.

\subsubsection{Interaction time}
It takes about 2 minutes to open a single articulated object, which is in the same order of magnitude as~\cite{xu2022UMP,eisner2022flowbot3d}, that take about 1 minute\footnote{These times were estimated from the demo videos on the project websites of both papers.}. Most of this time is spent in the actual robot motions, where we move slowly to keep the admittance controller stable and avoid shaky motions. The joint estimations take about 2s each (resulting in about 10 seconds spent estimating joint parameters). With additional efforts, the admittance controller could probably be tuned better to reduce the time needed for opening the objects but it will probably never be as fast as a force-controlled robot. And even so, it will still be much slower than the time it takes for humans to open articulated objects, which is in the order of seconds.
\subsubsection{fixed handle grasps}
We also found that using handle grasps that remain fixed during execution causes additional issues, which are not limited to or caused by proprioceptive information. On the one hand, the requirement for such grasps limits the workspace of the robot unnecessarily (see figure \ref{fig:examples-c}). On the other hand, it can result in collisions with the environment (see figure \ref{fig:examples-d}).
The first failure case could be solved by simply using larger or mobile robots. 
The second is more problematic. Think for example about dishwashers, which usually open towards the floor. We argue that future systems should be capable of breaking contact to regrasp the moving part based on the robot kinematics, joint parameters and the environment, and should maybe even deliberately use slip to change the grasp pose. Interesting work in this direction is~\cite{schiavi2022agent-aware-affordances}, where the authors use agent-aware affordances to determine where to grasp and when to regrasp for opening and closing articulated objects with given geometry and joint parameters.

    
     
\subsection{Should we use the proprioceptive information?}
\label{section:to-use-or-not}
In section \ref{section:analysis} we discussed that the baseline system we created is already quite capable of opening articulated objects, in line with the findings of Jain and Kemp~\cite{jain2010pulling}. By replacing their hook for a more generic parallel gripper, slip starts to become an issue, however. To make the proprioceptive information more useful, this should be tackled. One approach could be to design more suitable fingertips, although we believe the goal is to open the objects with a general-purpose system: opening cabinets is after all a means, not an end by itself. Alternatively one could attempt to model the slip or measure the slip using e.g. tactile sensors or visual odometry, which enables canceling the slip in the controller and/or taking it into account during joint estimation. This will make the system more complex though. Another issue is that slip is hard to capture in simulation, as friction is difficult to simulate realistically. This limits the ability to validate or train proprioception-based systems in simulation.
Furthermore, vision is still required for determining appropriate grasp poses. It could also speed up the system or make it more robust to make an initial estimation of the joint parameters before interacting, as in~\cite{heppert2023carto}. This brings us to the central question this paper wants to bring up: does proprioceptive sensing provide enough additional information when combined with vision to warrant the added complexity? Or can we manage with vision alone?


\subsection{Suggestions for evaluation of articulated object manipulation systems}
\label{section:discussion-evaluation}
In this section, we make some suggestions for evaluating articulated object manipulation systems. Based on  the findings in section~\ref{section:analysis}, we believe that evaluation protocols and/or benchmarks should incorporate the following aspects:
\begin{itemize}
    \item To evaluate systems that use proprioception in simulation, we have to attempt to provide realistic contacts to reduce sim2real gaps. Current simulation environments such as the UMP environment, join the gripper (or suction cup) with the cabinet through an artificial spring-like constraint~\cite{xu2022UMP, eisner2022flowbot3d}. This is perfect to incorporate some of the controller's compliance without handling the complex contact dynamics, but it does not suffice to properly evaluate systems that make use of contact information.
    \item The time needed to open the articulated objects should be optimized as well as the success rates and both should be reported to allow for a complete comparison between different systems.
    \item We should add appropriate collision objects in our evaluation. Articulated objects are not floating in a vacuum and this brings additional challenges, as discussed before. These challenges should be reflected in our evaluations. This was already mentioned in previous work, such as by Jain and Kemp~\cite{jain2010pulling}.
    \item We should incorporate opening/closing mechanisms, locking mechanisms and other joint dynamics. Many articulated objects (microwaves, drawers) have a locking mechanism that requires a certain amount of force to open. Other objects such as washing machines can have a handle that needs to be pressed to open. Yet other drawers have push-to-open, soft-close mechanisms, etc. Many cabinet doors also have a spring-like mechanism to close unless a certain opening angle is reached. This diversity should be represented in our evaluation. 

\end{itemize}

\section{CONCLUSIONS}
We combined previous work to enable a position-controlled robot equipped with a general-purpose parallel gripper to open articulated objects using proprioception. The success of such a system hinges on the degree to which the transform between the gripper and the handle remains fixed over time, i.e., the amount of slip that occurs during contact. Although slip occurs, our system was able to open several different articulated objects. Overcoming slip or simulating it to benchmark different systems creates additional complexity, which raises the question of whether we should reintroduce proprioception and fully embrace contact or if we can manage with vision-only systems and do not need to introduce additional complexity.




\section*{ACKNOWLEDGMENTS}
The authors wish to thank Nick Heppert, author of~\cite{heppert2022category} for open-sourcing the code used to estimate joint parameters and for the interesting discussions on articulated object manipulation.
The authors also wish to thank Cristian C. Beltran-Hernandez, author of~\cite{beltran2020learningforcecontrol}, and Frederik Ostyn for sharing their experience and expertise with compliant control. This research is supported by the Research Foundation Flanders (FWO) under grant number 1S56022N and the euROBIn Project (EU grant number 101070596).

\newpage
\bibliographystyle{IEEEtran}
\bibliography{references.bib}
\end{document}